\documentclass[11pt,a4paper]{article}
\usepackage[hyperref]{AACL-IJCNLP2020}
\usepackage{times}
\usepackage{latexsym}
\usepackage{graphicx}
\usepackage{xspace}
\usepackage{multirow}
\usepackage[utf8]{inputenc}

\DeclareUnicodeCharacter{2212}{\textendash}
\usepackage{microtype}

\aclfinalcopy 

\title{BERTChem-DDI : Improved Drug-Drug Interaction Prediction from text using Chemical Structure Information}

\author{Ishani Mondal \\
  Microsoft Research Lab \\
  Lavelle Road, Bengaluru, India \\
  \texttt{ishani340@gmail.com}}

\begin{document}
\maketitle
\begin{abstract}
Traditional biomedical version of embeddings obtained from pre-trained language models have recently shown state-of-the-art results for relation extraction (RE) tasks in the medical domain. In this paper, we explore how to incorporate domain knowledge, available in the form of molecular structure of drugs, for predicting Drug-Drug Interaction from textual corpus. We propose a method, BERTChem-DDI, to efficiently combine drug embeddings obtained from the rich chemical structure of drugs along with off-the-shelf domain-specific BioBERT embedding-based RE architecture. Experiments conducted on the DDIExtraction 2013 corpus clearly indicate that this strategy improves other strong baselines architectures by 3.4\% macro F1-score.
\end{abstract}

\section{Introduction}
Concurrent administration of two or more drugs to a patient to cure an ailment might lead to positive or negative reaction (side-effect). These kinds of interactions are termed as Drug-Drug Interactions (DDIs). Predicting drug-drug interactions (DDI) is a complex task as it requires to understand the mechanism of action of two interacting drugs. A large number of efforts by the researchers have been witnessed in terms of automatic extraction of DDIs from the textual corpus \cite{SAHU201815}, \cite{cnn}, \cite{Sun_2019}, \cite{li-ji-2019-syntax} and predicting unknown DDI from the Knowledge Graph \cite{8941958}, \cite{10.1145/3307339.3342161}. Automatic extraction of DDI from texts aids in maintaining the databases with high coverage and help the medical experts in their diagnosis and novel experiments.

In parallel to the progress of DDI extraction from the textual corpus, some efforts have been observed recently where the researchers came up with various strategies of augmenting chemical structure information of the drugs \cite{asada-etal-2018-enhancing} and textual description of the drugs \cite{PMID:32454243} to improve Drug-Drug Interaction prediction performance from corpus and Knowledge Graphs. 

The DDI Prediction from the textual corpus has been framed by the earlier researchers as relation classification problem. Earlier methods \cite{SAHU201815}, \cite{cnn}, \cite{Sun_2019}, \cite{li-ji-2019-syntax} for relation classification are based on CNN or RNN based Neural Networks. 

\begin{figure*}[!t]
\fbox{\includegraphics[height=9cm, width=0.9\textwidth]{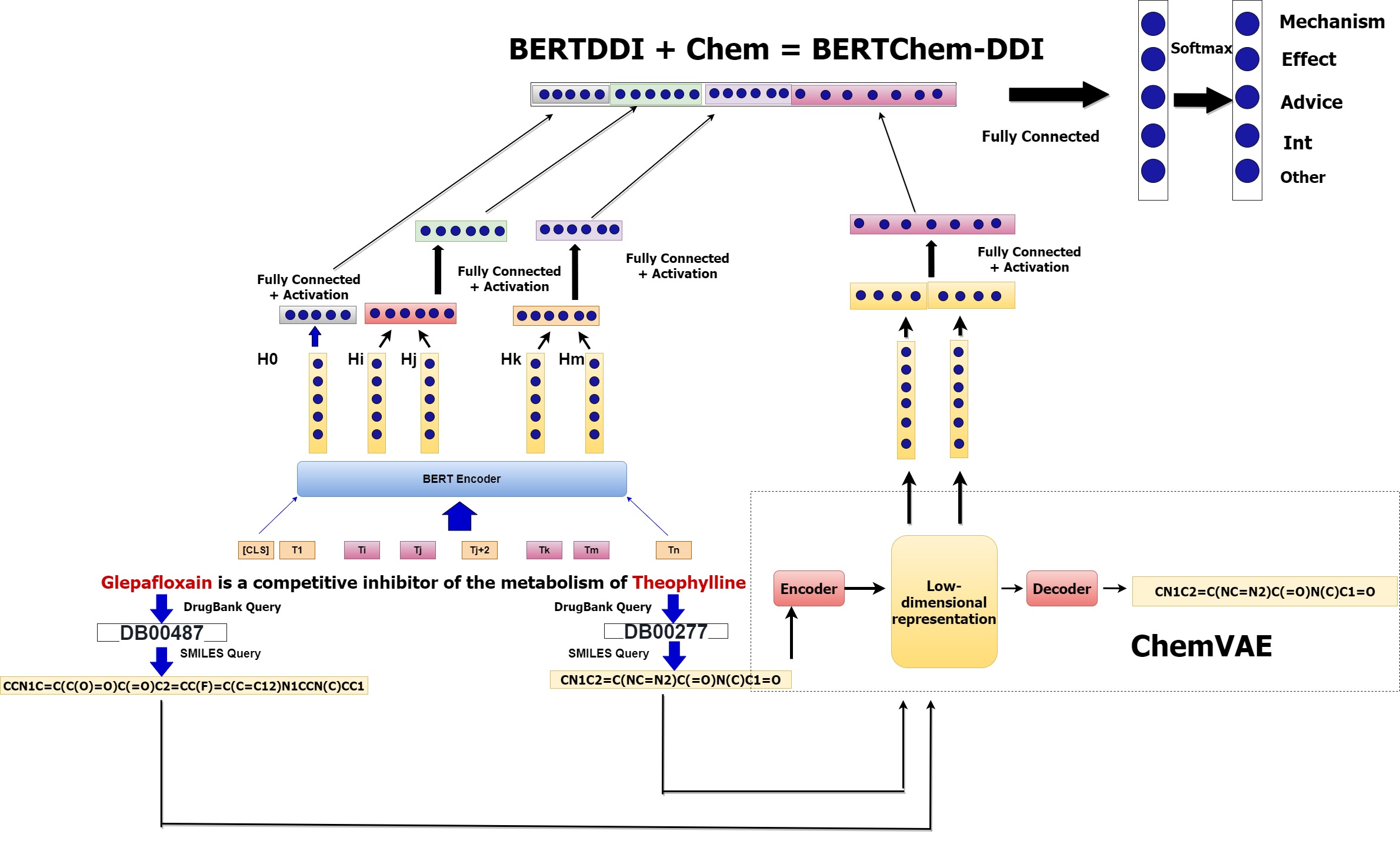}}
\centering
\caption{Schematic Representation of \emph{BERTChem-DDI} with the input sentence \textit{``Glepafloxain is a competitive inhibitor of the metabolism of Theophylline"} tagged with two drug entities \textbf{Glepafloxacin} and \textbf{Theophylline}.}
\label{fig-3}
\end{figure*}

Recently, with the massive success of the pre-trained language models \cite{Devlin2019BERTPO}, \cite{Yang2019XLNetGA} in many NLP classification / sequence labeling tasks, we formulate the problem of DDI classification as a relation classification task by leveraging both the entities and sentence-level information. We propose a model that leverages both domain-specific contextual embeddings (Bio-BERT) \cite{biobert} from the target entities and also external Chemical Structure information of the target entities (drugs). In the recent years, representation learning has played a pivotal role in solving various machine learning tasks. In addition to information of drug entities from the text, we make use of the rich hidden representation obtained from the molecule generation using Variational Auto-Encoder \cite{gomez} representation of the drugs to learn the chemical structure representation. During unsupervised learning of chemical structure information of the drugs using Variational AutoEncoder \cite{kingma2014autoencoding}, we make use of the canonical SMILES representation (\textbf{S}implified \textbf{M}olecular \textbf{I}nput \textbf{L}ine \textbf{E}ntry \textbf{S}ystem) obtained from the DrugBank \cite{drugbank}. We illustrate the overview of the proposed method in Figure 1. Experiments conducted on the DDIExtraction 2013 corpus \cite{HERREROZAZO2013914} reveals that this method outperforms the existing baseline models and is in line with the new direction of research of fusing various information to boost DDI classification performance.

In a nutshell, the major contributions of this work are summarized as follows:
\begin{itemize}
    \item We propose a method that jointly leverages textual and external Knowledge information to classify relation type between the drug pairs mentioned in the text.
    
    \item We show the molecular information from the SMILES encoding using Variational AutoEncoder helps in extracting DDIs from texts.
    
    \item Our method achieves new state-of-the-art performance on DDI Extraction 2013 corpus.
\end{itemize}

\section{Methodology}
Given a sentence $s$ with target drug entities $d_1$ and $d_2$, the task is to classify the type of relation ($y$) the drugs hold between them, $y$ $\in$ ($y_1$ , ...., $y_N$), where $N$ denotes the number of relation types. 

\subsection{Text-based Relation Classification}

Our model for extracting DDIs from texts is based on the pre-trained BERT-based relation classification model by Roberta. Given a sentence $s$ with drugs $d_1$ and $d_2$, let the final hidden state output from BERT module is $H$. Let the vectors $H_i$ to $H_j$ are the final hidden state vectors from BERT for entity $d_1$, and $H_k$ to $H_m$ the final hidden state vectors from BERT for entity $d_2$. An average operation is applied to obtain the vector representation for each of the drug entities. An activation operation 

\begin{equation}
    H_{1}^{'} = W_1 [tanh(\frac{1}{(j-i+1)}\sum_{t=i}^{j} H_t] + b_1
\end{equation}
\begin{equation}
    H_{2}^{'} = W_2 [tanh(\frac{1}{(m-k+1)}\sum_{t=k}^{m} H_t] + b_2
\end{equation}

\noindent
We make $W_1$ and $W_2$, $b_1$ and $b_2$ share the same parameters. In other words, we set $W_1$ = $W_2$ and keep $b_1$ = $b_2$. For the final hidden state vector of the first token (‘[CLS]’), we also add an activation operation and a fully connected layer, which is formally expressed as:

\begin{equation}
    H_{0}^{'} = W_0 (tanh(H_0)) + b_0
\end{equation}
Matrices $W_0$, $W_1$, $W_2$ have the same dimensions, i.e. $W_0$ $\in$ $R^{d * d}$ ,$W_1$ $\in$  $R^{d * d}$, $W_2$ $\in$  $R^{d * d}$, where $d$ is the hidden state size from BERT.

\noindent
We concatenate $H_{0}^{'}$, $H_{1}^{'}$ and $H_{2}^{'}$ and then add a fully connected layer and a softmax layer, which can be expressed as :

\begin{equation}
    h^{''} = W_3 [concat(H_{0}^{'}, H_{1}^{'}, H_{2}^{'})] + b_3
\end{equation}
\begin{equation}
    y_t^{'} = softmax(h^{''})
\end{equation}
 
\noindent
$W_3$ $\in$ $R ^ {N*3d}$, and $y_t^{'}$ is the softmax probability output over $N$. In Equations (1), (2), (3), (4) the bias vectors are $b_0$, $b_1$, $b_2$, $b_3$. We use cross entropy as the loss function. We denote this text-based architecture as \emph{BERT-DDI}.

\subsection{Chemical Structure Representation}
For the purpose of constructing an encoder from which a continuous latent representation is obtained, molecular representation of drugs has been used as both input and output. 

Gómez-Bombarelli et al. \citep{gomez} converted the discrete SMILES representations of the drug molecules into a continuous multi-dimensional representation using the unsupervised deep learning algorithm Variational Auto-Encoder(VAE) \cite{kingma2014autoencoding}. This representation has also been leveraged by \cite{8941958}. The input $x$ = ($x_1$ ,$x_2$ ,....,$x_n$ ) to VAE is represented by $x_i$ $\in$ $X$ where $X$ = {$C$, $=$, $($, $)$, $O$, $F$, $1$, $2$, · · · $9$} in the SMILES representation. Each $x_i$ is a X-dimensional one-hot vector. We denote this VAE architecture used in our experiments as \emph{ChemVAE} and is explained as follows: As an encoder it uses three 1D convolutional layers, followed by a single fully-connected layer. The decoder uses three layers of GRU networks. The objective of this work is to maximize the probability distribution of generation of SMILES representation of drug molecules with the help of latent representation as presented in equation below:
\begin{equation}
    P(X_{SMILES}) = \int P(X_{SMILES}\vert z) P(z) dz
\end{equation}

In equation 6, $X_{SMILES}$ denotes the drug molecules, $z$ represents the latent SMILES representation, $P(X_{SMILES})$ denotes the probability distribution of drug molecules. The \emph{ChemVAE} model takes SMILES representation of the drugs as input and encodes the drugs into continuous latent representation (z). The decoder then samples a string from the probability distribution over characters in the input SMILES representation. Finally, the hidden representation for each of the drug entities is treated as its chemical structure representation from \emph{ChemVAE}.

\begin{table}[!t]
\begin{center}
 \begin{tabular}{|c| c | c |} 
\hline
& Train  & Test \\ [0.5ex] 
 \hline
 No. of unique drugs & 2931 & 1055 \\
 No. of Normalized drugs & 2670 & 997 \\
 \hline
 No. of DDI Pairs &  27779 & 5713 \\
 \hline
 \end{tabular}
\end{center}
\caption{Statistics of the DDI Extraction corpus 2013. }
\label{stats}
\end{table}

\begin{table}[!t]
\small
\begin{center}
\begin{tabular}{|c|c|c|c|c|c|c|c|c|c|}
\hline
\multirow{2}{*}\textbf{Methods}& \textbf{Adv}&\textbf{Eff}&\textbf{Mch}&\textbf{Int}&\textbf{Tot}\\ \cline{2-6}
& \textbf{F1}  & \textbf{F1} & \textbf{F1} & \textbf{F1} & \textbf{F1}\\ \hline
\cite{SAHU201815} & 79 & 67 & 76 & 43 & 71 \\

Asada et al \cite{asada-etal-2018-enhancing} & 81 & 71 & 73 & 45 & 72 \\
\cite{zhang}  & 80 & 71 & 74 & 54 & 72 \\
\cite{Sun_2019}  & 80 & 73 & 78  & \textbf{58} & 75 \\
\cite{zheng}  & 85 & 76 &  77  & 57 & 77 \\
\cite{ZHU2020103451} & 86 & 80 & 84 & 56 & 80\\
\hline
Our method & \textbf{88} & \textbf{80} & \textbf{87} &  58 & \textbf{83} \\
\hline
\end{tabular}
\end{center}
\caption{Comparison of F1 scores for all the relation types using existing baselines on test set. \emph{Adv} indicates 'Advice', \emph{Mch} denotes 'Mechanism', 'Eff' means 'Effect', 'Tot' means overall.}
\label{baselines}
\end{table}

\subsection{BERTChem-DDI}
From the sentence $s$ containing two target drug entities $d_1$ and $d_2$, we obtain the chemical structure representation of two drugs $c_1$ and $c_2$ respectively using \emph{ChemVAE}. We concatenate these two embeddings $c_1$ and $c_2$ and pass those through a fully connected layer as represented as follows:

\begin{equation}
    chm = W [concat(c1, c2)] + b
\end{equation}

\noindent
$W$ and $b$ are the parameters of the fully-connected layer of the chemical structure representation of $d_1$ and $d_2$. The final layer of \emph{BERTChem-DDI model} contains the concatenation of all the previous text-based outputs (see Section 2.1) and chemical structure representation as expressed in the equations:

\begin{equation}
    o^{'} = W_3 [concat(H_{0}^{'}, H_{1}^{'}, H_{2}^{'}, chm)] + b_3
\end{equation}
\begin{equation}
    y_t^{'} = softmax(o^{'})
\end{equation}

\noindent
Finally the training optimization is achieved using the cross-entropy loss ($L_t$) : 

\begin{sloppypar} 
\begin{table}[!t]
\footnotesize
\begin{center}
\begin{tabular}{|c|c|c|}
\hline
\textbf{Models} & \textbf{Embeddings} & \textbf{Macro F1}\\ 
\hline
\emph{BERT-DDI} & \emph{bb v1.0 pubmed pmc} &  0.818 \\
\emph{BERTChem-DDI} & \emph{bb v1.0 pubmed pmc} &  0.829 \\
\emph{BERT-DDI} & \emph{bb v1.1 pubmed} & 0.822 \\
\emph{BERTChem-DDI} & \emph{bb v1.1 pubmed} & 0.838 \\
\hline
\end{tabular}
\end{center}
\caption{Probing deeper into the influence of chemical structure information into the BERT-based models for DDI Relation Classification. }
\label{bertchem-ddi}
\end{table}
\end{sloppypar}

\section{Experimental Setup}
In this section, we explain the dataset and experiments of using \emph{ChemVAE} and \emph{BERTChem-DDI}.

\subsection{Dataset and pre-processing}
We have followed the task setting of Task 9.2 in the
DDIExtraction 2013 shared task \cite{HERREROZAZO2013914} for the evaluation. This data set comprises of documents annotated with drug mentions and five types of interactions: \emph{Mechanism}, \emph{Effect}, \emph{Advice}, \emph{Int} and \emph{Other}. The task is a multi-class classification to classify each of the drug pairs in the sentences into one of the types and we evaluate using Precision (P), Recall (R) and F1-score (F1) for each relation type.

During pre-processing, we obtain the \emph{DRUG} mentions in the corpus and map those into unique \emph{DrugBank} identifiers. This mention normalization has been performed based on the longest overlap of drug mentions in the DrugBank. This mention normalization has been done for obtaining the corresponding \emph{SMILES} representation to encode molecular structure information. The dataset statistics of the total drugs and the normalized drugs are enumerated in table \ref{stats}. We initialize the non-normalized drug representations using pre-trained word2vec trained on PubMED \footnote{http://evexdb.org/pmresources/ngrams/PubMed/}.

\subsection{Training Details}
\begin{sloppypar}
We make use of the pre-trained contextual embeddings such as \emph{\textbf{bert-base-cased}}, \emph{\textbf{scibert-scivocab-uncased}} \cite{Beltagy2019SciBERTAP} and domain-specific \emph{\textbf{bb v1.0 pubmed pmc}} and \emph{\textbf{bb v1.0 pubmed}} as the initialization of the transformer encoder in \textbf{BERTChem-DDI}. We uniformly keep the maximum sequence length as 300, batch size 16, initial learning rate for ADAM optimizer as $2e-5$, drop out 0.1 for all the embedding ablations and trained for 5 epochs.
During unsupervised training of \emph{ChemVAE} with drugs from ZINC \cite{zinc}, the input SMILES representation has been trimmed to 120. The hidden dimension of \emph{ChemVAE} encoder is 200 and for the decoder it is 500. Finally, a 292-dimensional representation of the drugs has been ultimately used for initialization of the \emph{BERTChem-DDI} model's chemical structure representations of the drugs.
\end{sloppypar}

\begin{table}[!t]
\footnotesize
\begin{center}
\begin{tabular}{|c|c|}
\hline
\textbf{Embeddings on BERT-DDI} & \textbf{Test set Macro F1}\\ 
\hline
\emph{bert-base-cased} & 0.806 \\
\emph{scibert-scivocab-uncased} & 0.812\\
\emph{biobert v1.0 pubmed pmc} &  0.818 \\
\emph{biobert v1.1 pubmed} & 0.822 \\
\hline
\end{tabular}
\end{center}
\caption{Ablation of the contextual embeddings.}
\label{bertddi-ablation}
\end{table}

\section{Results and Discussion}
In this section, we provide a detailed analysis of the various results and findings that we have observed during experimentation. We have demonstrated strong empirical results based on the proposed approach for both text and chemical structure. We further want to understand the specific contributions by the chemical structure component besides the pre-trained BERT and its other domain-specific variants. For this purpose, we refer to our experimental configurations in meaningful ways while enumerating the results.
\paragraph{Ablation of Embeddings on BERT-DDI:}
During ablation analysis, we observe that the incorporation of domain-specific information in \emph{biobert v.1 pubmed} boosts up the predictive performance in terms of macro-F1 score (across all relation types) by 2.3\% compared to \emph{bert-base-cased}. Moreover, the \emph{scibert-vocab-cased} embedddings due to the scientific details obtained during fine-tuning achieves reasonable boost in performance. \emph{biobert v.1 pubmed based BERT-DDI} is thus the best-performing text-based relation classification model. The results are enumerated in Table \ref{bertddi-ablation}.

\paragraph{Advantage of Chemical Structure embeddings on BERTChem-DDI:}
During empirical analysis of the \emph{BERTChem-DDI} model, we observe how much performance gain can be achieved by augmenting the chemical structure information. From the results enumerated in terms of macro F1-score on all the relation types in table \ref{bertchem-ddi}, we observe that the best-performing \emph{BERT-DDI} model achieves a performance boost of 1.6\% after adding chemical structure information in \emph{BERTChem-DDI}. Probing deeper, we observe that the relation types \emph{Mechanism} (3.2\%) and \emph{Advice} (2.11\%) achieve significant performance improvement over \emph{BERT-DDI}.

\paragraph{Comparison with the existing baselines: }
We compare our best-performing model with some of the best-performing existing baselines. Our method achieves the state-of-the-art performance based on the results in Table.

\section{Conclusion}
In this paper, we develop an approach for DDI relation
classification based on pre-trained language model and chemical structure representation of drugs. Experiments on the benchmark DDI dataset proves the efficacy of our method. Possible directions of further research might be to explore Knowledge Graph based drug representation combined with textual description and other relation specific embeddings obtained from various ontologies.

\section*{Acknowledgement}{This work is an extension of the thesis work by the author during her course at the Indian Institute of Technology, Kharagpur. Besides, the author would also like to thank the anonymous reviewers for their insightful comments and feedback on the paper.}

\bibliography{aacl-ijcnlp2020}
\bibliographystyle{aacl-ijcnlp2020} 

\end{document}